\documentclass[10pt, a4paper]{article}
\usepackage{lrec2022} 
\usepackage{multibib}
\usepackage{multirow}
\newcites{languageresource}{Language Resources}
\usepackage{graphicx}
\usepackage{tabularx}
\usepackage{multirow}
\usepackage{soul}

\usepackage{titlesec}
\titleformat{\section}{\normalfont\large\bfseries\center}{\thesection.}{1em}{}
\titleformat{\subsection}{\normalfont\SmallTitleFont\bfseries\raggedright}{\thesubsection.}{1em}{}
\titleformat{\subsubsection}{\normalfont\normalsize\bfseries\raggedright}{\thesubsubsection.}{1em}{}
\renewcommand\thesection{\arabic{section}}
\renewcommand\thesubsection{\thesection.\arabic{subsection}}
\renewcommand\thesubsubsection{\thesubsection.\arabic{subsubsection}}

\usepackage{epstopdf}
\usepackage[utf8]{inputenc}
\usepackage{tipa}
\usepackage{hyperref}
\usepackage{xstring}
\usepackage{float}

\usepackage{color}
\usepackage{url}

\title{The Makerere Radio Speech Corpus: A Luganda Radio Corpus for Automatic Speech Recognition}
 
\name{Jonathan Mukiibi, Andrew Katumba, Joyce Nakatumba-Nabende, Ali Hussein, Josh Meyer} 

\address{Makerere University, Ronin Institute, Coqui \\
         Uganda, Egypt, USA \\
         jonathan.mukiibi@students.mak.ac.ug,\{andrew.katumba, joyce.nabende\}@mak.ac.ug,\\ alizawahry1@gmail.com, josh@coqui.ai\\}

\abstract{
Building a usable radio monitoring automatic speech recognition (ASR) system is a challenging task for 
under-resourced languages and yet this is paramount in societies where radio is the main medium of public communication and discussions. 
Initial efforts by the United Nations in Uganda have proved how understanding the perceptions of rural people who are excluded from social media is important in national planning.
However, these efforts are being challenged by the absence of transcribed speech datasets. In this paper, The Makerere Artificial Intelligence research lab releases a Luganda radio speech corpus of 155 hours. To our knowledge, this is the first publicly available radio dataset in sub-Saharan Africa. 
The paper describes the development of the voice corpus and presents baseline Luganda ASR performance results using Coqui STT toolkit, an open source speech recognition toolkit.  
 \\ \newline \Keywords{speech corpus, Luganda radio, automatic speech recognition } }

\begin{document}

\maketitleabstract

\section{Introduction}

In sub-Saharan Africa, low internet penetration makes radio the most preferred medium of social communication. Radio provides an opportunity for people's concerns, particularly in rural communities, to get heard through the various radio talk shows where they can call in. Uganda has over 309 licensed radio stations, which creates a unique platform where views that could potentially harness the development of better policies are discussed \cite{bbc}. Therefore, there is a need to retrieve such valuable perceptions for national development. 

Previous work in the area of radio browsing using Automatic Speech Recognition (ASR) has been done by the United Nations \cite{menon2018automatic}. They have also experimented with Keyword Spotting (KWS) systems in Uganda, and Somalia \cite{menon2018fast}. 
KWS for radio monitoring was developed as an alternative to ASR systems due to the lack of a large corpus of transcribed radio data. In this case, the conventional approach of using ASR to perform speech-to-text and then search through the lattices for the presence or absence of these keywords is not possible. 

In the last decade, the increase in the availability of large open-source speech datasets has propelled the application of deep learning in speech recognition research. As a result, research using various state-of-art ASR systems \cite{hannun2014deep} \cite{li2020espnet} has produced better results compared to the traditional machine learning approaches. However, the data demands of deep learning are well documented. Research on neural speech research for under-resourced languages is affected by the absence of speech datasets. On the other hand, this also frustrates the efforts to develop and adopt speech technologies in sub-Saharan Africa.

Our target language in this research is Luganda, which is a Bantu language spoken in the African Great Lakes region by more than fifteen million people \cite{uganda2016national}. Luganda faces the absence of publicly available speech and text resources like other low-resourced languages in sub-Saharan Africa. Currently, there are no open-source Luganda speech datasets that are available. To fill this gap, Makerere AI lab\footnote{\url{https://www.air.ug}} in partnership with Mozilla, has made efforts to add Luganda as a language on the Common Voice platform\footnote{\url{https://commonvoice.mozilla.org/lg}}. However, the Common Voice dataset \cite{ardila2019common} is different compared to a radio dataset. Building ASR models for radio requires a radio-specific dataset. Such a dataset should be able to capture unique radio settings such as background noise, telephone speech, studio speech, news reports, and adverts. In this paper, \textit{we collect speech and text data, as well as using transfer learning, an approach that is optimized for under-resourced training}. We use the openly available Kinywarwanda ASR model \cite{meyer2019multi} and fine-tune the checkpoints to use the collected Luganda Common Voice dataset and radio corpus.

The main contributions of this paper are: 
\begin{enumerate}
    \item We present the methodology used to collect and create the Luganda radio speech corpus.
    \item We openly release 155 hours of the radio dataset. 
    \item We present the first radio monitoring Connectionist Temporal Classification (CTC) end-to-end ASR model for Luganda using transfer learning with 203 hours of read speech data and 120.7 hours of radio data.
    \item We evaluate the performance of the ASR model on a COVID-19 radio conversation test set to establish the model's effectiveness in monitoring COVID-19 related keywords. 
    \item We show how hotword boosting can improve keyword detection in a COVID-19 use case-based radio monitoring system and evaluate the model's performance on gender.

\end{enumerate}
The remainder of the paper is organized as follows: In Section \ref{relatedwork}, we discuss related work in ASR concerning the datasets used, then we discuss our corpus development approach in section \ref{corpusdev}. In Section \ref{lugandamodel}, we present the Luganda Automatic Speech Recognition model. Section \ref{modelevaluation} discusses the model performance and evaluation. Finally, Section \ref{conclusion} concludes the paper.

\section{Related Work}\label{relatedwork}

This section reviews related work in Automatic Speech Recognition systems for radio monitoring and approaches taken in corpus creation. Previous work with radio data has proven valuable for plant disease monitoring, and prediction \cite{akera2019keyword} using a keyword spotter model. In this case, radio monitoring using Keyword Spotting System (KWS) model was used together with the Adhoc mobile surveillance approach \cite{mutembesa2018crowdsourcing} to replace traditional surveying methods. 
Efforts have been made to develop KWS solutions for under-resourced languages for radio monitoring. Work has been carried to develop quickly deployable systems for ASR-free keyword spotting approaches. The system uses a multilingual bottleneck feature extractor trained on well-resourced out-of-domain languages \cite{menon2018feature}. The aim of this work was to support United Nations humanitarian relief efforts by using radio data in parts of Africa with severely under-resourced languages. \cite{menon2018fast} proposes a KWS radio browsing system that uses dynamic time warping (DTW) as supervision for training a convolutional neural network (CNN) based keyword spotting system using a small set of spoken isolated keywords. 

Research by \cite{menon2017radio} \cite{saeb2017very} \cite{menon2018automatic} has been done in using machine learning for radio monitoring. \cite{menon2017radio} presents the initial efforts of extracting information from broadcast radio speech in Uganda for Ugandan English, Acholi, and Luganda. The ASR monitoring system uses Hidden Markov Model (HMM), Gaussian Mixture Model (GMM), Subspace Gaussian mixture Model (SGMM), and Deep Neural Network (DNN) based acoustic models as keyword spotters \cite{menon2017radio}. They used  a train set of 9 hours and a 62 min test set resulting into a 52.47\% best word error rate (WER) with SGMM-BMMI and 53.54\% word error rate with a DNN and HMM models. \cite{saeb2017very} also presents a radio browsing system developed on a tiny corpus of annotated speech by using supervised training of multilingual DNN and HMM acoustic models. The research in \cite{saeb2017very} presents interesting examples of using radio for humanitarian monitoring by carrying out different pilots on various topics discussed on the radio like natural disasters, refugees, health service delivery, and malaria. \cite{menon2018automatic} also presents initial efforts in developing an ASR system for Somali using 1.57 hrs of annotated radio speech data. The research by \cite{menon2018automatic} uses a combination of CNNs, Time-delay Neural Networks (TDNNs), and Bi-directional Long Short Term Memory (BLSTMs) to achieve a WER of 53.75\%.

The related work discussed in this section presents applications of radio monitoring. However, this work does not mention any publication of open radio datasets. Furthermore, the research uses traditional approaches and KWS that can manage to feed off small annotated datasets. In this paper, we collect read speech and radio speech to mitigate challenges with limited data. This enables us to experiment with deep learning approaches that have led to significant improvements in word error rates.

The advent of Deep Learning toolkits like Mozilla's DeepSpeech, which is based on Baidu's Deep Speech \cite{hannun2014deep} has recently been improved as Coqui STT\footnote{\url{https://github.com/coqui-ai/stt}}. Other toolkits like SpeechBrain \cite{speechbrain}, NVIDIA NeMo \cite{kuchaiev2019nemo} are a result of increased research in end-to-end speech recognition. Recent research in speech for African languages by  \cite{dossou2021okwugb} presents OkwuGbé, an end-to-end approach for building ASR systems for low resourced African languages with the case study of Igbo and Fon.

Coqui STT has presented a higher performance at higher efficiency for various languages \cite{tyers2021shall}. Coqui STT has been tested for both research and production. Recent research with Coqui STT has produced good results for the German \cite{agarwal2019german} and English languages. A WER of 21.5\% is presented for German on a combination of Tuda, Voxforge, and Mozilla datasets \cite{agarwal2019german}. A WER of 4.5\% for English\footnote{\url{https://coqui.ai/english/coqui/v1.0.0-huge-vocab}} on the Librispeech clean dataset\footnote{\url{https://www.openslr.org/12}}.

We use the Coqui STT toolkit to develop a Luganda model using 203 hours of read speech data from the Common Voice dataset and 120.7 hours of transcribed radio speech data.

\section{Corpus Development}\label{corpusdev}

The following section outlines the development of the Makerere Radio Speech Corpus which we release under a Creative Commons license, as well as other corpora (e.g. Common Voice) used during the training and testing phases of experimentation. Statistics for the Makerere Radio Speech Corpus can be found in Table~\ref{releaseddataset}

\begin{table}[H]
\begin{center}
 \begin{tabular}{|l|c|c|} 
 \hline
 & \textbf{Gender} & \textbf{Duration (hrs)}\\
 \hline
 Transcribed & --- & 20 \\
 \hline
 \multirow{3}{*}{Untranscribed} & Women & 1.4 \\
 & Men & 4.6\\
 &--- & 129 \\
 \hline
 \textbf{Total}&  & \textbf{155} \\
 \hline
\end{tabular}
\caption{Release statistics for the Makerere Radio Speech Corpus. Number of hours are reported across gender where known. Most data we present in this release of the corpus is untranscribed, but still has gone through multiple filtering steps to ensure it is high-quality (e.g. not broadcast music, split on pauses, etc.)}
\label{releaseddataset}
\end{center}
\end{table} 

First we will discuss the creation of the Makerere Radio Speech Corpus and then our use of Common Voice. 

\subsection{Makerere Radio Speech Corpus}

Summary statistics for the Makerere Radio Speech Corpus can be found in Table~\ref{releaseddataset}. 

\subsubsection{Radio Data Collection}

We collected radio data by recording streams from online Luganda radio stations. We did this daily from 06:00 to 23:00 for a minimum period of three months for over ten radio stations. The priority for which times to record was based on the public radio live broadcasting schedules.

\subsubsection{Radio Data Transcription}
After audio recording, the next step was transcription. The transcription process follows precise rules around the transcriber writing all the words they hear with exceptional cases on the numbers, titles, dates, and punctuation. All numbers have to be written as words, titles (e.g. Luganda equivalents of "Mrs." and "Dr.") have to be written out in full just as they sound in speech, dates and times are written out in the way they were spoken, and no punctuation was used. Transcription is a very resource-intensive process, and it becomes more challenging with radio data. Radio data is characterized by background noise/music, overlapping speech, filler pauses, breaths, incomplete or partial words, telephone speech, and unintelligible speech. We worked around this by using an automated data selection criteria and creating transcription guidelines that the transcribers followed. The guidelines defined how all the posed challenges and edge cases had to be transcribed by Luganda linguists. 

\begin{figure}[!h]
\begin{center}
\includegraphics[width=0.5\textwidth]{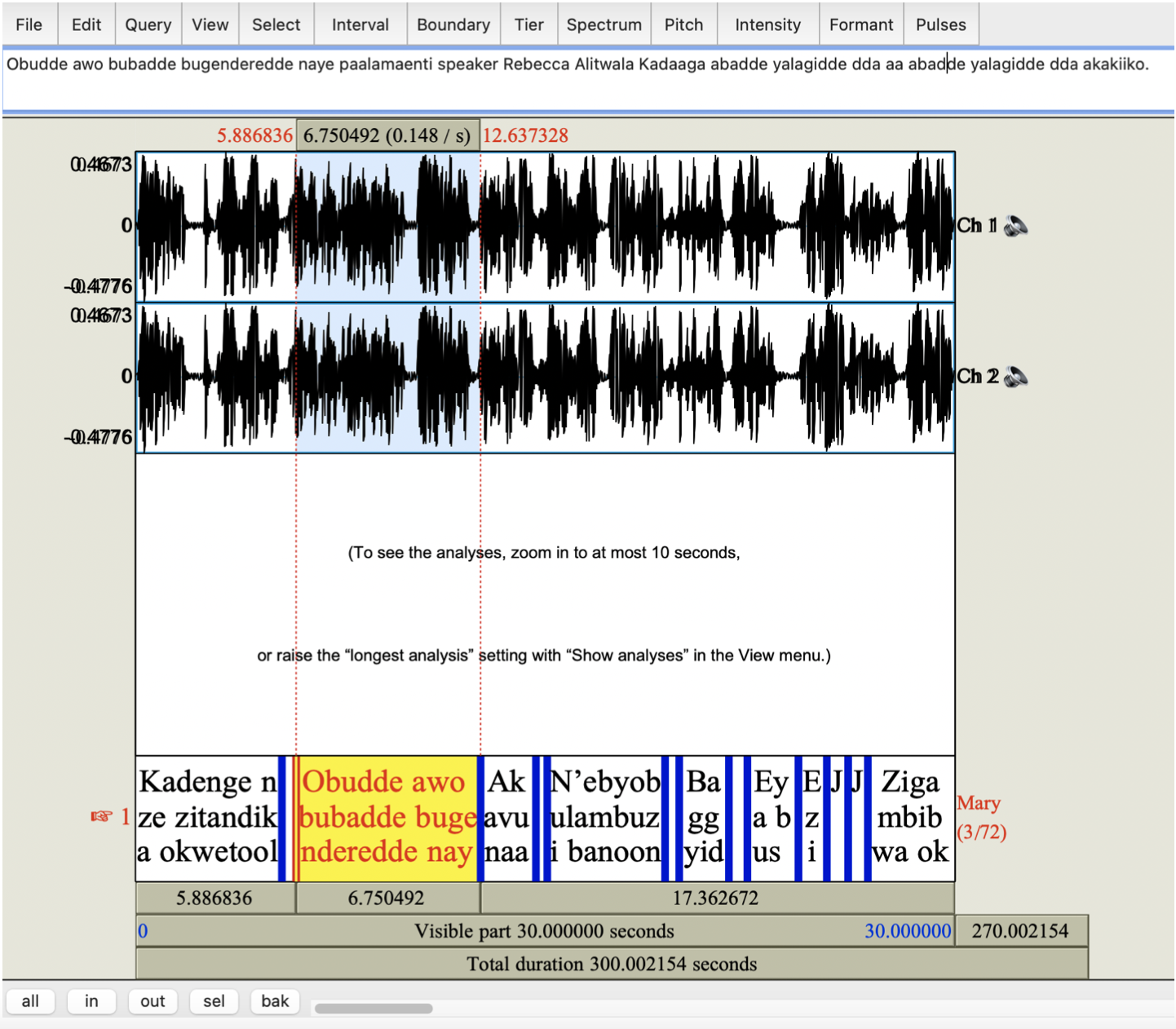}
\caption{Data transcription using the Praat tool.}
\label{fig:transcribe}
\end{center}
\end{figure}

We developed an audio selection tool\footnote{\url{https://github.com/AI-Lab-Makerere/COVID-19-ASR}} based on py-webrtcvad \cite{sredojev2015webrtc} and DeepSpeech\footnote{\url{https://github.com/mozilla/DeepSpeech}} to automatically identify sections of audio that are likely to have human speech. We randomly sampled audio transcriptions from every transcriber to calculate the transcription WER. We obtained a WER of 0.3\%. The radio data was transcribed using the Praat annotation tool\footnote{\url{https://www.fon.hum.uva.nl/praat/}} as shown in Figure \ref{fig:transcribe}.

\subsubsection{Radio Data Preparation}
The transcribed audio data was saved in MP3 format. The audio files were saved along with the transcript in Textgrid file format. The audio files were then converted to WAV file, mono-channel with a sampling rate of 16kHZ, and the results saved to a CSV file. The transcripts were cleaned to remove all known encoding errors and extra-linguistic tags like ``um", and ``laughter", which were added as part of the transcription guidelines. During the process of exporting transcripts, encoding errors were observed. These encoding errors resulted from foreign words and names, where diacritics interacted with vowels. These were changed to the base vowel (e.g. ``ö" was replaced by ``o"). Table \ref{table:csv} shows an example of a sample CSV file.

\begin{table}[!h]
\begin{center}
 \begin{tabular}{|l|c|p{2.5cm}|  }

 \hline
  \textbf{wav\textunderscore filename} & \textbf{wav\textunderscore filesize} & \textbf{transcript}\\ [0.3ex] 
 \hline
 \small audio\textunderscore 5fbb5.wav & \small 316844 & \scriptsize{kwegamba ensigo zino} \\
 \hline
 \small audio\textunderscore 5fb42.wav & \small188204 & \scriptsize{osobola okugamba nti}\\
 \hline
 \small audio\textunderscore 5fbb5.wav & \small 201644 & \scriptsize{wekatandika okukula} \\
 \hline
 \end{tabular}
 \caption{Sample metadata for a cleaned and filtered dataset.}
 \label{table:csv}
 \end{center}
\end{table}

The initial radio dataset of 95.0 hours was split into 82.7 hours for training, 10.5 hours for validation and 1.8 hours for testing. The testing set was obtained from a radio station which was not part of both the training and validation set. 
We carried out a listening exercise using 5 people. These listened to 82.7 hours of training data and established the gender of speaker(s)'s voice(s) in the audio file. Table \ref{gendertrain} shows the number of hours in the Training, Validation and Testing sets. It also shows the number of hours of women and men's voices in the training set.
\begin{table}[H]
\begin{center}
 \begin{tabular}{|l|c|} 
 \hline
 \textbf{Dataset} & \textbf{Hours}\\
 \hline
 Training &  82.7\\
 \hline
 Validation & 10.5\\
 \hline
 Testing & 1.8\\
 \hline
  \textbf{Total} & 95.0\\
 \hline
\end{tabular}
\caption{Statistics for the dataset used in the first round of training. We have statistics on gender representation for the training set. In number of hours of training data, we had: women (7.5), men (67.8), and audio with multiple speakers where there were both men and women speaking (7.4).}
\label{gendertrain}
\end{center}
\end{table}

In addition to the 95.0 hours described in Table \ref{gendertrain}, we transcribed more 25.7 hours radio data. These were combined together to obtain 120.7 hours of transcribed radio data. Table \ref{finalradio} shows the number of hours, word tokens and word types in the final radio dataset. We used the 1.8 hours transcribed from a radio station which is not part of the training and validation as the test set. We then split the remaining 118.9 hours into 90\% training and 10\% validation set.

\begin{table}[H]
\begin{center}
 \begin{tabular}{|l|c|c|c|} 
 \hline
  & \textbf{Tokens} & \textbf{Types} & \textbf{Hours}\\
 \hline
 Training & 900,608  & 135,647 & 107.1\\
 \hline
 Validation & 99,839  & 27,939 & 11.8\\
 \hline
 Testing & 14,117 & 5,110 & 1.8 \\
 \hline
  \textbf{Total} & --- & --- & 120.7 \\
 \hline
\end{tabular}
\caption{Statistics for the transcribed radio dataset used to train the Luganda radio ASR. Shown are word types, word tokens, and hours of audio. The Makerere Radio Speech Corpus includes a 20 hour subset of the data shown in this table, where we received permission from the radio station to release the subset under a Creative Commons license.}
\label{finalradio}
\end{center}
\end{table}

\subsubsection{Open Radio Data Corpus}

We release a corpus of 155 hours publicly available online under the Creative Commons BY-NC-ND 4.0 license and can be downloaded from Zenodo\footnote{\url{https://doi.org/10.5281/zenodo.5855017}}. The dataset release comprises of:
\begin{enumerate}
    \item 20 hours of human transcribed radio speech. The audio is sampled at 16kHZ, mono-channel.
    \item Two CSV files for the 20-hour human transcribed dataset - cleaned.csv contains cleaned transcripts and uncleaned.csv contains uncleaned transcripts. The uncleaned transcripts contain extra speech details included in tags like [laughter] for laughter, and [um] for filler pauses, which speaker is talking, where each speaker is assigned an identifier A or B.
    \item A transcription guideline.
    \item A multi-speaker untranscribed dataset of 6 hours of radio data. 1.4 hours of women voices and 4.6 hours of men voices. Each audio is a ten-seconds clip with a single speaker. 
    \item 135 hours of multi-speaker untranscribed radio data.
\end{enumerate}
The 20 hours of human transcribed radio dataset were used in our experiments. The rest of the dataset was not used. 

\subsection{Common Voice Dataset}

Common Voice  is a crowdsourcing project started by Mozilla to create a free database for speech recognition software \cite{ardila2019common}. It is a platform\footnote{\url{https://commonvoice.mozilla.org/lg}} where anyone can donate their voice to an open-source data bank\footnote{\url{https://commonvoice.mozilla.org/lg/datasets}}.
We collected 300 hours of Luganda voice on the Common Voice platform. The Common Voice dataset has each entry consisting of a unique MP3 file and a corresponding text file. Part of the recorded hours in the dataset also include demographic data like age, and gender. The Luganda Common Voice dataset was contributed by 39.2\% women and 33.5\% men while the remaining 27.3\% were anonymous contributors.
Table \ref{cv_table} shows a detailed breakdown of the Luganda Common Voice dataset based on the age of the contributors.
\vspace{-2mm}
\begin{table}[H]
\begin{center}

\begin{tabular}{ |p{2cm}|c|  }
 \hline
 \textbf{Age} & \textbf{Percentage (\%)} \\
 \hline
 19-29 & 41.4 \\
 30-39 & 21.1 \\
 40-49 & 5.8 \\
 50-59 & 3.0 \\
 Unclassified & 28.7 \\
 \hline
 \textbf{Total} & 100.0\\
  \hline
\end{tabular}
\caption{Luganda Common Voice (CV) corpus demographics. CV dataset was used together with radio data to train the Luganda ASR model}
\label{cv_table}
\end{center}
\end{table}

\subsubsection{Common Voice Data Preparation}
The Common Voice dataset is released with a clips folder, invalidated.tsv, reported.tsv, train.tsv, dev.tsv, other.tsv, validated.tsv and test.tsv files. The dataset splits are done by the Mozilla's CorporaCreator\footnote{\url{https://github.com/mozilla/CorporaCreator}} in the form of 80\% train, 10\% validation and 10\% test sets. The dataset contains MP3 audio files. The proposed speech recognition toolkit expects the audio files to be in WAV format, mono-channel, and with a 16kHz sampling rate. Using the Common Voice importer python script, the Common Voice data was processed to comma-separated values (CSV) files (train, dev and test) and the audio files were converted to WAV format. The CSV file has the format of (wav\textunderscore filename,wav\textunderscore filesize,transcript). The wav\_filesize in bytes is used to group audio of similar lengths for efficient batching. We used the English alphabet as our output alphabet. We used the commonvoice-utils\footnote{\url{https://github.com/ftyers/commonvoice-utils}} package to perform basic linguistic checks to identify characters that were not defined in the alphabet. We only used 203 hours out of 300 hours of Common Voice data because the remaining hours were not validated. Table \ref{cvtokens}, shows the number of word tokens, types and hours in the Common Voice dataset used for training. 

\begin{table}[H]
\begin{center}
 \begin{tabular}{|l|c|c|c|} 
 \hline
 \textbf{Dataset} & \textbf{Tokens} & \textbf{Types}  & \textbf{Hours}\\
 \hline
 Training & 414,129  & 74,340 & 162.4\\
 \hline
 Validation & 92,969  & 25,040 & 20.3\\
 \hline
 Testing & 92,708 &  24,700 & 20.3 \\
 \hline
 \textbf{Total} & --- & --- & 203.0 \\
 \hline
\end{tabular}
\caption{Statistics for the Luganda Common Voice (CV) dataset. Shown are word types, word tokens, and hours of audio.}
\label{cvtokens}
\end{center}
\end{table}

\section{Luganda Automatic Speech Recognition Model} \label{lugandamodel}

The section presents the Luganda ASR model trained and evaluated on the radio dataset. We describe the model architecture, the training process, and the language model. 

\subsection{Model Architecture}
\begin{figure}[h]
\includegraphics[width=0.5\textwidth]{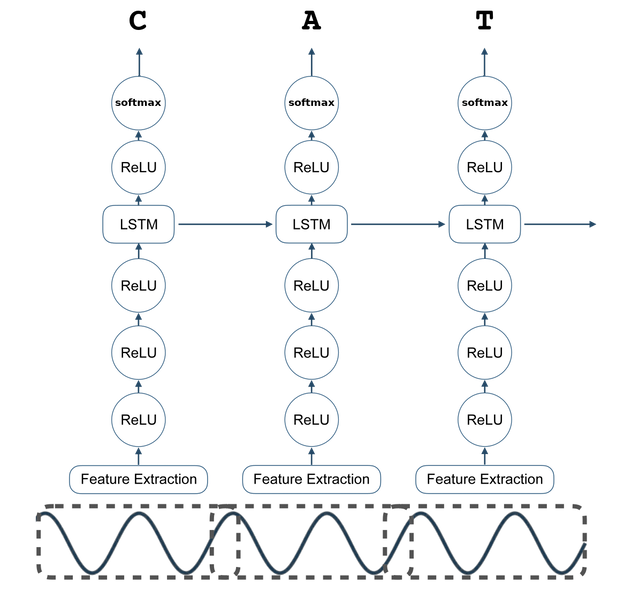}
\caption{Coqui STT architecture}
\label{fig:coquiengine}
\end{figure}

The Luganda ASR model is a Coqui Speech-to-Text (STT) model.  Coqui STT's architecture\footnote{\url{https://stt.readthedocs.io/en/latest/Architecture.html}} is based on Baidu's Deep Speech research \cite{hannun2014deep}. However, further improvements have been made, and the core of the engine is now of recurrent neural network (RNN) trained to ingest speech spectrograms and generate text transcriptions\footnote{\url{https://stt.readthedocs.io/en/latest/}} (see Figure \ref{fig:coquiengine}).

Coqui STT uses a probabilistic algorithm called Connectionist temporal classification (CTC)\cite{hannun2017sequence}.  An algorithm commonly used to train deep neural networks. The algorithm aligns input sequences of audio and output sequences of characters.

The model architecture is setup as follows. Let a single utterance $x$ and label $y$ be sampled from a training set:
\[ S= \{ {(x^{(1)},y^{(1)}), (x^{(2)},y^{(2)}),...} \}  \]

Each utterance, \(x^{(i)}\) is a time-series of length \(T^{(i)}\) where every time-slice is a vector of audio features, \({x_t}^{(i)}\) where \(t=1,…,T^{(i)}\). 

Mel-frequency cepstral coefficients (MFCC) are used as the features whereby \({x_{t,p}}^{(i)}\) denotes the $p^{th}$  MFCC feature in the audio frame at time $t$. The purpose of the Recurrent Neural Network (RNN) is to convert an input sequence into a sequence of character probabilities for the transcription, with \({\hat{y_t}}=\mathbf{P}{({{c_t}|x})}_t\), where for Luganda $c_t \in$ $\{a,b,c,...,z,space,apostrophe,blank\}$. The Connectionist Temporal Classification (CTC) loss uses blank to indicate transitions between characters.

The RNN model has five layers of hidden units. Consider a given input \({x}\), the hidden units at layer \({l}\) are denoted with the convention that \(h^{(0)}\) is the input. The first three layers are not recurrent. For the first layer, at each time \({t}\), the output depends on the MFCC frame \({x_t}\) along with a context of \({C}\) frames on each side. We use \({C} =9\) for our experiments. The remaining non-recurrent layers operate on independent data for each time step. Thus, for each time \({t}\), the first three layers are computed by:

\[{h_t}^{(l)}={g(W^{l}{h_t}^{(l-1)}+{b}^{(l)})}\]

where $g(z)=min\{max\{0,z\},20\}$ is a clipped rectified-linear (ReLu) activation function and $\{W^{(l)}$, ${b}^{(l)}\}$ are the weight matrix and bias parameters for layer \({l}\). The fourth layer is a recurrent layer. The layer includes a set of hidden units with forward recurrence \({h}^{(f)}\) as:

\[{h_t}^{(f)}={g(W^{(4)}{h_t}^{(3)}+ {W_r}^{(f)}{h_{t-1}}^{(f)}+{b^{(4)}})})\]

Note that \({h}^{(f)}\) must be computed sequentially from $t=1$ to \({t=T^{(i)}}\) for the $i^{th}$ utterance. The fifth (non-recurrent) layer takes the forward units as inputs:
\[{h}^{(5)}={g(W^{(5)}{h}^{(f)}+{b}^{(5)})}\]

The output layer is standard logits that correspond to the predicted character probabilities for each time slice \({t}\) and character \({k}\) of the alphabet:

\[{h_{t,k}}^{(6)} = {\hat{y_{t,k}}}={(W^{(6)}{h_t}^{(5)})_k + {(b_k)}^{(6)}}\]

Here \({b_k}^{(6)}\) denotes the $k^{th}$ bias and \({({W}^{(6)}{h_t}^{(5)})}_k\) the $k^{th}$ element of the matrix product. Once we have computed a prediction for \({\hat{y}}_{t,k'}\), we then compute the CTC loss \({L}{({\hat{y}},{y})}\) to measure the error in prediction. The CTC loss requires the above to indicate transitions between characters. During training, we can evaluate the gradient \({\nabla}{L}{({\hat{y}},{y})}\) with respect to the network outputs given the ground-truth character sequence \({y}\). From this point, computing the gradient with respect to all of the model parameters may be done via back-propagation through the rest of the network. We used the Adam method for training.

\subsection{Model Training}

We utilized a cross-lingual transfer learning approach \cite{meyer2019multi} to get a good performing model. We used a two-tier pre-training approach for transfer learning.  We chose to transfer learn from a pre-trained English DeepSpeech model to Kinyarwanda. Kinyarwanda and Luganda are linguistically related. They are both tonal Bantu languages that have. 
This can be expressed at different language dimensions: Phonetically, both languages are tonal \cite{jerro2018linguistic}, and syntactically some words have got similar meanings, for example: ``abantu" meaning ``people" or ``humans", ``akantu" meaning ``little thing". Morphologically, they have some similar noun classes. They also follow the same grammatical principles for one noun class (singular) to shift into another noun class to give the plural of that noun class. Table \ref{noun} shows two examples where Kinywarwanda and Luganda show similarities in noun classes. 
\begin{table}[H]
\begin{center}
\begin{tabular}{|p{0.8cm}|p{1.2cm}|p{2.2cm}|p{2cm}|}
 \hline
 \textbf{Class} & \textbf{Number} &\textbf{Kiywarwanda} & \textbf{Luganda} \\
 \hline
 1 & Singular&umu- (umuntu) & (o)mu- (omuntu) \\
 \hline
 2 & Plural& aba- (abantu) & (a)ba- (abantu)\\
 \hline
\end{tabular}
\caption{Luganda noun class morphology.}
\label{noun}
\end{center}
\end{table}


We trained the model using both the Luganda radio and the Luganda Common Voice datasets described in section \ref{corpusdev}. 

First, a pre-trained English release model was downloaded and fine-tuned using Kinywarwanda Common Voice data. The English model checkpoints were fine-tuned multiple times, first to Kinyarwanda, then ultimately to Luganda. 
We used the English alphabet across all the languages to ease the fine-tuning process. For Luganda, we replace all occurences of ``\textipa{ŋŋ}" with ``ng" so that all text characters in the training data correspond to the English alphabet. We trained the pre-trained English model for 200 epochs to get a Kinyarwanda model. We then fine-tuned the Kinyarwanda model to Luganda for 200 epochs. 

We performed two training rounds. In the first training round, we used the radio dataset described in Table \ref{gendertrain}. It has 82.7 hours of training data, 10.3 hours of the validation data and 1.8-hours held out test set. In the second round of training, we used the final radio dataset described in Table \ref{finalradio} and the Common Voice dataset described in Table \ref{cvtokens}.
The training was done with 107.1 hours of training, 11.8 hours of validation and 1.8-hours of test data from radio. We also combined this with 162.4 hours of training, 20.3 hours of validation and 20.3 hours of test data from Common Voice. 
For the hyperparameters, we use a dropout of 0.1 with batchsize of 64 for training and validation and a learning rate of 0.0001. We also performed time and frequency mask augmentation during training. 

\subsubsection{Luganda Language Model}
 
We used a probabilistic language model to build a scorer for our acoustic Luganda model. Using a Kenlm toolkit \cite{heafield2011kenlm}, we build a 3-gram Language Model(LM). We used a text corpus of 80,000 sentences for the first language model and initial tests. We then use a text corpus of 500,000 sentences for the second round of tests. The Luganda sentences were extracted from online news Luganda websites, PDF documents from authors, and the Luganda Bible. The corpus was cleaned with one sentence per line. Table \ref{lmcorpus} shows the counts for sentences, word tokens and word types in each text corpus used to build the language model. 

\begin{table}[H]
\begin{center}
 \begin{tabular}{|p{2.7cm}|p{2cm}|p{1.8cm}|} 
 \hline
 \textbf{Language Model} & \textbf{Word Tokens} & \textbf{Word Types} \\
 \hline
 80,000 sentences & 972,104 & 151,281 \\
 \hline
 500,000 sentences & 6,682,657  & 609,755 \\
 \hline
\end{tabular}
\caption{Number of sentences, word types and tokens in each text corpus that was used to build the language model.}
\label{lmcorpus}
\end{center}
\end{table}

\section{Model Performance and Evaluation}\label{modelevaluation}

We trained the Luganda ASR model with 82.7 hours of radio data as shown in Table \ref{gendertrain} and obtained a WER of 65.1\% on the radio test set of 1.8 hours. In this case, we use a text corpus with 80,000 sentences to create the language model. The details of the corpus are provided in Table \ref{lmcorpus}.

In the second round of training, the training set was three times of that used in the first round of training.  We also increased the number of sentences in the text corpus from 80,000 to 500,000. The details of the text corpora are showed in Table \ref{lmcorpus}. In Table \ref{table:wer}, we present much better results for the Luganda ASR model during this training phase. The WER was calculated on both the 1.8 hours radio test set and 20.3 hours Common Voice test set.  

\begin{table}[h]
\begin{center}
\begin{tabular}{|l|c| }
\hline
 \textbf{Dataset} & \textbf{WER (\%)} \\  
\hline
 Common Voice & 33 \\
 \hline
 Radio & 47 \\
 \hline
\end{tabular}
\caption{WER on Common Voice and Radio dataset.}
\label{table:wer}
\end{center}
\end{table}

The model performs better on Common Voice data because this is read speech data containing one speaker for every audio clip and less background noise. As earlier discussed, radio data is conversational and has unique characteristics which explains the differences in the WER on both datasets.

\subsection{Hotword boosting}
 Using Coqui STT's hotwords weighting feature, we biased the predictions of our Luganda ASR model on selected keywords using the hotword boosting technique. The algorithm ``boosts" the likelihood of a selected hotword. During decoding, the language model assigns likelihoods to words as they are recognized. The boost is an additive factor to the language model's original likelihood. It makes the keyword more likely in the beam search. We performed these tests using the model after the first round of training. When a negative boost value is applied, there are chances that homophones might be used instead in case they exist in the audio. As a result, the behaviour of the keywords of interest was observed by adjusting different boost values to obtain the best boost values for a given keyword. This was also analysed to understand how the boost values affected specific keywords that are homophones in nature.
 
While boosting the hotwords, we used a range of -1000 to +1000. The boost values for each keyword were determined by assigning a boost range of values from -1000 to + 1000 for each keyword. We then observed the model results on applying values between -1000 to +1000. In the case where the keyword was boosted, we recorded the boost value.
We logged the results of the different boost values in the range to understand which boost values worked best. 
Table \ref{table:hotword} shows an example of hotword boosting for the word ``ekifo" by changing it to ``ekifuba" ("cough" in English) which was the word mentioned in the audio file. The change happened at the +200 to +1000.0 boost values. The original transcript obtained with STT model was: ``eno oba ne virus eno eyitibwa covid okolola ebifuba enayumba abantu balina okwegendereza ekifo tulina gugaawulira e emabegako".

\begin{table*}[ht]
\begin{center}
\begin{tabular}{|c|c|c|  }
 \hline
 \textbf{Boost Value} & \textbf{Transcript} & \textbf{Verdict}\\
 \hline
 -1000& ... abantu balina okwegendereza \textbf{ekifo} tulina gugaawulira ... & false negative\\
 \hline
-600 & ... abantu balina okwegendereza \textbf{ekifo} tulina gugaawulira ... & false negative\\
 \hline
-200& ... abantu balina okwegendereza \textbf{ekifo} tulina gugaawulira ... & false negative\\
 \hline
  0 & ... abantu balina okwegendereza \textbf{ekifo} tulina gugaawulira ... & false negative\\
 \hline
 +200 & ... abantu balina okwegendereza \textbf{ekifuba} e u o i na gugaawulira ... & true positive\\
 \hline
 +600 & ... abantu balina okwegendereza \textbf{ekifuba} e u o i na gugaawulira ... & true positive\\
 \hline
 +1000 & ... abantu balina okwegendereza \textbf{ekifuba} e u o i na gugaawulira ... & true positive\\
 \hline
\end{tabular}
\caption{How the transcript changes with boosting the keyword ``ekifuba" using boost values of -1000.0, -600.0, -200.0, 0.0 200.0, 600.0 and 1000.0. A boost value of 0.0 effectively means that no boost was used.The keyword was mentioned in the audio but the Luganda ASR had failed to transcribe it properly. }
\label{table:hotword}
\end{center}
\end{table*}

\subsection{Comparison of ASR performance with hotword boosting}
In this section, we evaluate the performance of the Luganda ASR model with hotword boosting (HTWD-B) verses using ASR without hotword boosting (ASR). The evaluation was done on five prominent COVID-19 keywords from radio discussions. The purpose was to find out whether hotword boosting can assist in detecting COVID-19 keyword mentions which might be missed by the ASR model, in which case we may choose to run inference using the ASR model while boosting certain keywords for which the ASR model is under performing. 


We created two test datasets of 10 second audio clips.  The test dataset was created using new unseen audio. In the first dataset, each audio file was listened to by a linguist to confirm the presence of the keywords of interest. The dataset included the following mentions of keywords: 
\begin{itemize}
\item Eighty three (83) audio files contained \textbf{``covid"} or \textbf{``kovidi"} (English ``covid")
\item Sixteen (16) audio files contained \textbf{``ekirwadde"} (English ``disease").
\item Six (6) audio files contained \textbf{``kolona"} (English ``corona").
\item Five (5) audio files contained \textbf{``ssennyiga"} (English ``flu").
\item Two (2) audio files contained \textbf{``ekifuba"} (English ``cough").
\end{itemize}

For each audio file, we provided a boost range of -1000 to +1000 for the boost values and specified the keyword of interest. We then observed the keyword behaviour across 6 different boost values of -1000.0, -600.0, -200.0, 200.0, 600.0 and 1000.0. 

Table \ref{table:covidtest} shows the results based on the tests carried out with the five keywords. We observe that using ASR with hotword boosting returns more True Positive (TP) results. All the False Negatives that returned True Positives results did so at boost values of 200.0, 600.0 and 1000.0. However, both approaches perform well on the \textbf{``covid"} keyword.

\begin{table}[H]
\begin{center}
 \begin{tabular}{|p{2cm}|p{1cm}|p{1cm}|p{0.8cm}|p{0.6cm}|} 
 \hline
 \multirow{2}{*}{\textbf{Keyword}} & \multicolumn{2}{|c|}{\textbf{ASR}} & \multicolumn{2}{|c|}{\textbf{HTWD-B}} \\\cline{2-5}
 & TP & FN & TP & FN \\
 \hline
 ``covid" & 71 & 12 & 71 & 12 \\
 \hline
 ``ekirwadde" & 11 & 5 & 14 & 2 \\  
 \hline
 ``kolona" & 3 & 3 & 6 & 0 \\ 
 \hline
 ``ssennyiga" & 5 & 0 & 5 & 0 \\
 \hline
 ``ekifuba" & 1 & 1 & 2 & 0 \\
 \hline
 
 \end{tabular}
 \caption{True Positive (TP) and False Negative (FN) results on COVID-19 test set based on a Luganda ASR model with hotword boosting (HTWD-B) and the ASR without hotword boosting (ASR).}
 \label{table:covidtest}
\end{center}
\end{table}

In the second dataset, each audio was listened to by the linguist in order to confirm that the keywords of interest were absent. We collected 122 10-second random radio recordings as our test set in this dataset. The results in Table \ref{table:nocovid} show that boosting the word \textbf{``kolona"} results in 6 False Negatives out of 122 radio recorded files. 



\begin{table}[H]
\begin{center}
 \begin{tabular}{|p{2cm}|p{1cm}|p{1cm}|p{0.8cm}|p{0.6cm}|} 
 \hline
 \multirow{2}{*}{\textbf{Keyword}} & \multicolumn{2}{|c|}{\textbf{ASR}} & \multicolumn{2}{|c|}{\textbf{HTWD-B}} \\\cline{2-5}
 & FP & TN & FP & TN \\
 \hline
 ``covid" & 0 & 122 & 0 & 122 \\
 \hline
 ``ekirwadde" & 0 & 122 & 0 & 122 \\  
 \hline
 ``kolona" & 0 & 122 & 6 & 116 \\ 
 \hline
 ``ssennyiga" & 0 & 122 & 0 & 122 \\
 \hline
 ``ekifuba" & 0 & 122 & 0 & 122 \\
 \hline
 \end{tabular}
 \caption{False Positive and True Negative results on a non COVID-19 test set.}
 \label{table:nocovid}
\end{center}
\end{table}
\vspace{-3mm}
Based on the True positives (TP), False Positives (FP), False Negatives (FN), and True Negatives (TN), we calculated the precision and recall for both approaches to obtain F-score result of 0.94 with hotword boosting (HTWD-B). From the results shown in Table \ref{table:fscores}, we noticed minimal improvement in the F-score for HTWD-B. However, this still presents the potential of hotword boosting in under-performing ASR systems where a given use case is of priority.

\begin{table}[H]
\begin{center}
 \begin{tabular}{|p{3cm}|p{1.5cm}|p{1.5cm}|} 
 \hline
\textbf{Metric} & \textbf{ASR} & \textbf{HTWD-B} \\ 
 \hline
 True Positives & 91 & 98  \\
 \hline
 False Positives & 0 & 6 \\  
 \hline
 False Negatives & 21 & 14 \\ 
 \hline
 Precision & 1 & 0.99 \\ 
 \hline
 Recall & 0.81 & 0.89 \\ 
 \hline
 Fscore & 0.89 & 0.94 \\ 
 \hline
 \end{tabular}
 \caption{Fscore results for hotword boosting (HTWD-B) and ASR without hotword boosting (ASR).}
 \label{table:fscores}
 \end{center}
\end{table}

\subsection{Model Evaluation: Gender Bias}

Bias mitigation is a serious problem in Artificial Intelligence (AI) research. Over the past decade, academia has increased the amount of time its researchers have spent studying bias in machine learning models \cite{genderbias1,genderbias2}. Academic studies by researchers with diverse backgrounds have played a major role in preventing bias in AI models. Managing bias in speech recognition is a very important aspect especially when the speech technology is a solution that will be used by diverse users. As a first step to understand gender bias, several datasets like the Artie Bias Corpus \cite{meyer-etal-2020-artie} and the curated subset of the English Mozilla Common Voice corpus have been released for testing for gender bias.

We therefore carried out gender bias tests to get an understanding of how our Luganda ASR model generalises on women and men radio speech. We did this by creating a 28 minutes test set that contained 14 minutes of women's speech and 14 minutes of men's speech. The test dataset was selected from new unseen radio data sorted from radio studio discussions by linguists. The linguists manually listened to each audio file to ascertain the speaker. We tested the Luganda ASR model obtained after the first round of training. The results are shown in Table \ref{table:gender}.

\begin{table}[h]
\begin{center}
 \begin{tabular}{|p{1.8cm}|p{1cm}|p{2.4cm}| }
 \hline
  \textbf{Gender} & \textbf{WER} & \textbf{Duration (mins)}  \\ [0.5ex] 
 \hline
 Women & 70.6\%  & 14\\
 \hline
 Men & 53.5\% & 14\\ [1ex] 
 \hline
 \end{tabular}
 \caption{Model performance on a held-out gender test set.}
 \label{table:gender}
 \end{center}
\end{table}

As shown in Table \ref{table:gender}, our model is biased towards men's voices with a better WER of 53.5\% compared to the WER of 70.6\% for the women voices. This can be explained by the existing bias in the training data described in Table \ref{gendertrain} which was used to train the model. The training set contains 81.9\% of men voices, 9.0\% of women voices and 8.9\% of the discussions with both women and men voices. It is probable that the overall WER of the model can be improved significantly if the model is able to transcribe women’s voices better. It is apparent that reducing gender bias in the dataset leads to more effective and accurate models~\cite{meyer-etal-2020-artie}. We suggest that any speech data collection strategy should ensure that women’s and men's speech is equally represented if the model is to generalize well for real life scenarios. 

\section{Conclusion}\label{conclusion}

This paper presents a Luganda radio corpus and Luganda ASR for radio monitoring. We show how we utilized transfer learning to fine tune a Kinyarwanda model on Luganda Common Voice and radio data. We evaluate the performance of the Luganda ASR model on a held-out test set to obtain the best WER of 33\% on Common Voice and 47\% radio dataset. We evaluate the model’s performance on a held-out test-set of COVID-19 keywords to obtain Fscore of 0.94. We highlight the importance of gender consideration in ASR models by evaluating our model on women’s and men's voices.  We release the Makerere Radio Speech Corpus, a Luganda radio corpus of 155 hours. We believe that this work has the potential to benefit many researchers working on radio monitoring work in sub-Saharan Africa. 

\section{Acknowledgement}
This work is funded by Bill and Melinda's Gates Foundation OPP1212027 and a grant from the International Development Research Centre, Ottawa, Canada and the Swedish International Development Cooperation Agency. We are grateful to GIZ's Fair Forward and Mozilla for funding the data transcription process. Special thanks go to the management of Radio Simba, Tropical FM for availing us with radio data.

\section{Bibliographical References}\label{reference}

\bibliographystyle{lrec2022-bib}
\bibliography{lrec2022-example}

\label{lr:ref}
\bibliographystylelanguageresource{lrec2022-bib}

\end{document}